# Neural Architecture Search with an Efficient Multiobjective Evolutionary Framework

Maria G. Baldeon Calisto, Susana K. Lai-Yuen*
Department of Industrial and Management Systems Engineering, University of South Florida,
4202 E. Fowler Ave., Tampa, FL USA 33620

**ABSTRACT**

Deep learning methods have become very successful at solving many complex tasks such as image classification and segmentation, speech recognition and machine translation. Nevertheless, manually designing a neural network for a specific problem is very difficult and time-consuming due to the massive hyperparameter search space, long training times, and lack of technical guidelines for the hyperparameter selection. Moreover, most networks are highly complex, task specific and over-parametrized. Recently, multiobjective neural architecture search (NAS) methods have been proposed to automate the design of accurate and efficient architectures. However, they only optimize either the macro- or micro-structure of the architecture requiring the unset hyperparameters to be manually defined, and do not use the information produced during the optimization process to increase the efficiency of the search. In this work, we propose EMONAS, an **E**fficient **M**ulti**O**bjective **N**eural **A**rchitecture **S**earch framework for the automatic design of neural architectures while optimizing the network´s accuracy and size. EMONAS is composed of a search space that considers both the macro- and micro-structure of the architecture, and a surrogate-assisted multiobjective evolutionary based algorithm that efficiently searches for the best hyperparameters using a Random Forest surrogate and guiding selection probabilities. EMONAS is evaluated on the task of 3D cardiac segmentation from the MICCAI ACDC challenge, which is crucial for disease diagnosis, risk evaluation, and therapy decision. The architecture found with EMONAS is ranked within the top 10 submissions of the challenge in all evaluation metrics, performing better or comparable to other approaches while reducing the search time by more than 50% and having considerably fewer number of parameters.

## 1. INTRODUCTION

Deep learning methods have become very successful at solving a variety of complex tasks such as image classification and segmentation, speech recognition, and machine translation. However, the performance of a neural network is highly dependent on the configuration of its architecture and hyperparameters. Extensive work has focused on manually designing network configurations for the task in hand. As deep neural networks have greatly increased in size and complexity to achieve better performance, manually designing a neural network resembles a black-box optimization process that requires extensive experience, time, and computational resources. At the same time, recent work has shown that deep neural networks are usually over-parametrized and can be significantly reduced in size without a loss of accuracy [1]. Therefore, there is a growing interest on automating the design of accurate and efficient deep neural network architectures.

To address the aforementioned issues, neural architecture search (NAS) frameworks have been proposed to automate the design of neural network architectures. Optimization methods based on evolutionary algorithms [2, 3, 4] and reinforcement learning [5, 6, 7] have been presented to search for the best architectural hyperparameters. Although NAS algorithms can identify competitive architectures, the search process is usually time consuming and computationally expensive taking even hundreds or thousands of GPU days to obtain state-of-the-art architectures [2, 3, 5]. In an effort to reduce the computational load, works have proposed gradient based optimization [8, 9], weight sharing [10, 11] and one-shot architecture search [12]. However, these methods only optimize the network´s accuracy, and cannot be easily adapted to include additional objective functions or constraints in the design process.

*laiyuen@usf.edu

Recently, multiobjective NAS frameworks that consider multiple objective functions have been presented for image classification and language processing tasks [13, 14]. Nevertheless, limited works have been presented for the more complicated task of 3D medical image segmentation, which plays a crucial role in medical applications such as clinical diagnosis, computer-assisted surgery and treatment planning. This is due to the limited labeled datasets, and high dimensionality and variability of the 3D image data. Existing NAS frameworks design 2D segmentation architectures [15, 16, 17, 18] and 3D networks [19, 20, 21, 22]. However, 2D architectures do not fully exploit inter-slice information and can have suboptimal performance in some medical datasets. Meanwhile, the methods proposed for designing 3D architectures either search for the micro-structure (structure of the encoder-decoder cell), or search for the macro-structure (optimal number of cells and their connection). Therefore, manual engineering is still required to find the topology of the non-optimized structure. Furthermore, most of the networks are usually designed focusing only on reducing the segmentation error, do not consider the simultaneous optimization of other objective functions such as minimizing the size of the network, and do not address the efficiency of the search process.

In this work, we present EMONAS, an **E**fficient **M**ulti**O**bjective **NAS** framework for 3D image segmentation that searches for both accurate and efficient architectures. EMONAS is composed of a novel search space that includes the hyperparameters that define the micro- and macro-structure of the architecture, and a Surrogate-**a**ssisted **M**ultiobjective **E**volutionary based **A**lgorithm (SaMEA algorithm) that efficiently searches for the best hyperparameter values in the search space. The proposed SaMEA algorithm incorporates selection probabilities to guide the search to the most promising subproblems in the Pareto Front and improve the selection of hyperparameter values to mutate during evolution. Furthermore, a Random Forest surrogate model is introduced to approximate an architecture´s segmentation performance, and reduce the number of candidate architectures trained and search time. EMONAS is evaluated on the task of cardiac segmentation on the MICCAI ACDC challenge [23] and is ranked within the top 10 submissions of the leaderboard in all evaluation metrics, performing better or comparable to other AutoML and NAS methods while being smaller and requiring considerable less computational time for the architecture search.

The contributions of this work are threefold. First, we propose a novel search space that simultaneously considers the micro- and macro-structure of the architecture, which reduces the need for manual interventions because no prefixed template of the encoder-decoder cell or the depth/width of the macro architecture need to be defined beforehand. Secondly, a Surrogate-assisted Multiobjective Evolutionary based Algorithm (SaMEA) is presented to construct accurate and efficient segmentation architectures. The SaMEA algorithm takes advantage of the information generated during the initial generations of the evolutionary search to improve the exploration of the search space and convergence. This allows the framework to construct architectures in fraction of the time in comparison with other NAS methods proposed for 3D medical image segmentation. Furthermore, although in the current experiments we construct architectures that minimize both the size and segmentation error of the architecture, other objective functions can be easily incorporated for optimization such as inference time or energy consumption. Finally, we present the network configuration found with the EMONAS framework in the task of cardiac segmentation that performs better or similar to other top automatically designed architectures while being significantly smaller.

## 2. METHODS

The EMONAS framework is comprised of two key components: the micro and macro search space and the SaMEA search algorithm. The EMONAS architecture is automatically constructed for a specific dataset by using the SaMEA algorithm to search for the best micro and macro hyperparameters that minimize the expected segmentation error and size of the network.

## 2.1 Search Space

EMONAS searches for an encoder-decoder architecture with an equal number of encoder and decoder cells as shown in Figure 1. Each cell in the encoder path is followed by a max-pooling operation with a stride of 2 that halves the size of the input feature map. Meanwhile, in the decoder path a transpose convolution is applied to double the size of the feature map. The cells on the opposite sides of the encoder-decoder path are connected through a summation operation to promote information and gradient flow. The last convolutional layer in the architecture has a kernel of size 1x1x1 and a softmax activation function. The constructed networks have this fixed encoder-decoder structure to reduce the search space and improve the efficiency of the optimization process. The rest of the architectural hyperparameters are strategically summarized in just 10 decision variables in what we denominate the micro and macro search space, allowing us to further improve the search efficiency.

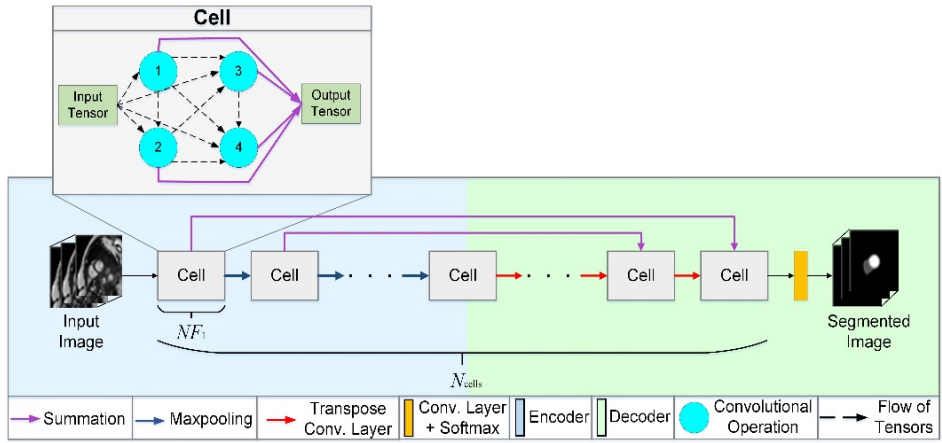

Figure 1. The encoder-decoder network architecture. The macro search space includes the hyperparameters that define the depth and width of the architecture. The micro search space includes the hyperparameters that define the configuration of the encoder-decoder cell.

### Macro Search Space
The macro search space includes the hyperparameters that determine the number of encoder-decoder cells in the architecture, the number of filters on each cell, and the learning rate that has to be adjusted when the architecture changes. This group of hyperparameters allows our algorithm to find the appropriate depth and width of the architecture, and optimize the number of parameters.

**Number of cells ($N_{cells}$):** The number of cells in the architecture is computed by $N_{cells} = 2n_c+1$, $n_c \in$ [2,3,4]. Where $n_c$ cells are assigned to the encoder path, $n_c$ cells to the decoder path and one cell connects the encoder and decoder paths.

**Number of filters on cell $i$ ($NF_i$):** We use the common heuristic of doubling the number of filters after a max-pooling operation and halving them after a transpose convolution. Therefore, the number of filters for each cell $i$ ($NF_i$) can be computed after finding the number of filters on the first encoder cell $NF_1$, where $NF_1=2^{n_f}$, $n_f \in$ [3,4,5]. $n_f$ is set to an integer value to make $NF_1$ be a power of 2, as commonly used in segmentation networks.

## Micro Search Space

The micro search space proposed in our work includes the hyperparameters that define the configuration of the encoder-decoder cell. The cell is represented by a directed acyclical graph with $B$ nodes. Each node represents a convolutional operation, while a directed edge represents the flow of tensors between operations. For each node $b \in B$ two hyperparameters are defined that are the input tensor to the node and the type of convolutional operation applied.

**Input tensor to node $b$ ($I_b$):** The set of possible input tensors $I_b$ to node $b$ include the set of output tensors from all previous nodes in the cell and the cell´s input tensor. For example in Figure 1, the set of possible inputs to node 3 are the cell's input tensor, the output tensor of node 1 and the output tensor of node 2.

**Type of convolutional operation in node $b$ ($O_b$):** Inspired by [9], the set of possible convolutional operations $O_b$ for node $b$ is composed of 2D, 3D and Pseudo-3D convolutions (P3D). Each convolutional layer is comprised of a ReLU activation function, the selected convolutional operation, and an instance normalization layer. This group of convolutional operations allows the NAS framework to select between the analysis of in-plane information captured by inexpensive 2D convolutions, volumetric information captured by costly 3D convolutions, and inter-slice and intra-slice information from anisotropic images from the P3D convolutions.

The number of nodes in the cell has been set to $B=4$ considering the work in [6, 8]. Finally, the output of the cell is the summation of the output tensor of all nodes to improve information gradient transmission. A summary of all the 10 decision variables being optimized and their corresponding search range are presented in Table 1.

Table 1. Hyperparameters being optimized by the SaMEA algorithm and their search range.

| Hyperparameter | Formula | Search Range |
| --- | --- | --- |
| Input tensor to node 2 ($I_2$) | - | [Input tensor, node 1] |
| Input tensor to node 3 ($I_3$) | - | [Input tensor, node 1, node 2] |
| Input tensor to node 4 ($I_4$) | - | [Input tensor, node 1, node 2, node 3] |
| Convolutional operation node 1 ($O_1$) | - | Refer to Section 2.1. |
| Convolutional operation node 2 ($O_2$) | - | Refer to Section 2.1. |
| Convolutional operation node 3 ($O_3$) | - | Refer to Section 2.1. |
| Convolutional operation node 4 ($O_4$) | - | Refer to Section 2.1. |
| Number of cells ($N_{cells}$) | $2n_c + 1$ | $n_c \in [2,3,4]$ |
| Number of filters for $NF_1$ | $2^{n_f}$ | $n_f \in [3,4,5]$ |
| Learning rate | - | $[1\times10^{-6}, 9\times10^{-6}]$ |

### 2.2 Multiobjective SAMEA Algorithm

The SaMEA algorithm searches for the best hyperparameter values $x$ in the search space $\Omega$ by solving the following optimization problem:

$$Min\ f_1(x) = \alpha\big(C - MCDice_{Train}(\theta)\big) + \big(C - MCDice_{Val}(\theta)\big) + \beta\left(\frac{E-e_{max}}{E}\right) \quad (1)$$

$$Min\ f_2(x) = \log(|\theta|) \quad (2)$$

$$subject\ to\ x\ \in \Omega \quad (3)$$

Where $f_1(x)$ is the expected segmentation error (ESE) function presented in [10], and measures a network´s segmentation error through the multi-class Dice coefficient in the training set ($MCDice_{Train}(\theta)$) and validation set ($MCDice_{Val}(\theta)$), and the distance between the total number of training epochs ($E$) and the epoch with the maximum validation multi-class Dice coefficient ($e_{max}$). $C$ is the number of segmentation classes. By using the multi-class Dice coefficient in the loss function, the cost of each class has a standardized value that ranges between 0 and 1 (a loss value near 0 indicates a significant overlap between the predicted segmentation and the ground truth, while values near 1 signify a small spatial overlap). Hence, all segmentation classes are given the same weight in the loss function and the tendency to be biased towards a specific segmentation class decreases. Meanwhile, $f_2(x)$ quantifies the logarithm of the number of parameters in the network, where $\theta$ are the parameters learned by the network and $|\cdot|$ is the cardinality operator. Solving this discrete non-convex hyperparameter optimization problem is challenging because the search space is large, evaluating the solutions is costly and there are no guarantees of optimality.

The proposed SaMEA algorithm is an evolutionary-based algorithm that approximates the set of solutions that provide the different quality trade-offs among the objective functions, known as Pareto Front. It is based in the MEA [16] and MOEA/D [24] algorithm, which have shown to closely approximate the true Pareto Front and produce a diverse set of solutions. However, instead of relying on random mutation to generate new candidate architectures, which can be inefficient at producing good solutions, a hyperparameter mutation probability is proposed to increase the probability of selecting hyperparameter values that have produced a good segmentation performance on previously tested architectures. For this objective, each hyperparameter value is scored using the ESE function. Let $S_{h_{ij},G}$ be the score for the hyperparameter $i$ with value $j$ at generation $G$, where $i$ refers to a hyperparameter (i.e., number of cells) and $j$ to the specific value assigned (i.e., 5, 7 or 9). Then, $S_{h_{ij},G}$ is computed as:

$$S_{h_{ij},G} = \sum_{g=1}^{G-1} \frac{I(f_g) * [ESE_{max} - ESE(f_g)]}{I(f_g)} \quad (4)$$

Where $ESE(f_g)$ is the ESE function for candidate architecture $f_g$ tested in generation $g$. $ESE_{max}$ is the maximum ESE value any architecture can obtain. Meanwhile, $I(f_g)$ is an indicator function that has a value of 1 if the hyperparameter $h_{ij}$ was applied to construct the candidate architecture $f_g$ and 0 otherwise. The probability of mutating to hyperparameter value $j$ at generation $G$ ($P_{h_{ij},G}$) is obtained as follows:

$$P_{h_{ij},G} = \frac{PS_{h_{ij},G}}{\sum_{j \epsilon J} PS_{h_{ij},G}} \quad (5)$$

where
$$PS_{h_{ij},G} = \frac{S_{h_{ij},G}}{\sum_{j \epsilon J} S_{h_{ij},G}} + \varepsilon \quad (6)$$

$J$ is the set that contains all the values of hyperparameter $i$. To make sure all $P_{h_{ij},G} > 0$, $PS_{h_{ij},G}$ is first computed where an $\varepsilon = 0.002$ value is added.

During evolution, the MEA algorithm solves each subproblem once in a generation. However, previous studies have shown that some subproblems discover more Pareto optimal solutions than others [12]. Therefore, to exploit the most promising search regions, subproblems that have actively contributed to approximate the Pareto Front or provide a diverse set of solutions will have a higher probability of being selected to be solved and can be solved more than once in a generation. For this aim, we introduce the subproblem selection probabilities similar to [14].

The most time-consuming step during the evolutionary search is training the candidate architectures. Therefore, we propose to incorporate a Random Forest surrogate model to approximate a network´s ESE value without the need of training it. Furthermore, to assess the level of uncertainty the Random Forest has in the prediction, we calculate the standard deviation of the individual trees´ predictions. This dispersion measure increases as the new point goes farther from the data available in the training set. Therefore, we use it to select the candidate architectures that will be trained during evolution to increases the fidelity of the prediction model. A Random Forest model is selected after finding it performs better in terms of the prediction´s mean square error, mean absolute error and $R^2$ score against a radial basis function (RBF) with cubic, linear and TPS kernel, a Gaussian Process with RBF kernel, a feedforward neural network, an extra random forest model and a multiple linear regression model. Moreover, Random Forest works well with categorical input data, can be trained with a limited number of samples, and has lower computational cost compared with commonly used predictors (i.e., Gaussian processes, neural networks).

---

**Algorithm 1:** SaMEA Algorithm

**Input:** Population size $N$, Neighborhood size $T$, Max generations $G$, Learning generations $LG$

1. **Initialization Phase**
1.1 Use LHS to generate the initial population $\{x_1^1…x_1^N\}$ of size $N$ from the search space $\Omega$.
1.2 Train the $N$ architectures and calculate the objective functions $F(x_1^j) = [f_1(x_1^j), f_2(x_1^j)] \; \forall \; j \in \{1,2 …, N\}$
1.3 Initialize *NDS*, and *STP*.
2. **Learning Phase**
2.1 **for** $g=2:LG$ **do**:
2.2     **for** $i=1:N$ **do**:
    **2.2.1** Select two solutions $x_g^k$ and $x_g^l$ from neighborhood of $i$. Create new solution $x_g^j$ by applying crossover and random mutation.
    **2.2.2** Train the architecture and calculate objective functions $F(x_g^j) = [f_1(x_g^j), f_2(x_g^j)]$.
    **2.2.3** Update *NDS, STP,* and *PNS* with PBI approach.
3. **Exploitation Phase**
3.1 **for** $g=LG:G$ **do**:
3.2     **for** $i=1:N$ **do**:
    **3.2.1** Select subproblem $n$ using probabilities from [14]. Choose two solutions $x_g^k$ and $x_g^l$ from the neighborhood of $n$. Create new solution $x_g^j$ by applying crossover and mutation according to probabilities from Eq. (5).
    **3.2.2** Train Random Forest and predict ESE value. Compute $\hat{F}(x_g^j) = [\hat{f}_1(x_g^j), f_2(x_g^j)]$.
    **3.2.3** Select architecture if it satisfies criteria in section 2.2. Otherwise discard and return to 3.2.1.
    **3.2.4** Train the architecture and calculate objective functions $F(x_g^j) = [f_1(x_g^j), f_2(x_g^j)]$.
    **3.2.3** Update *NDS, STP,* and *PNS*
**Return** the population of *NDS*

---

The SaMEA algorithm is presented in Algorithm 1. It is divided into 3 phases: 1) Initialization phase, 2) Learning phase, and 3) Exploitation phase.

**Initialization phase:** The algorithm is initialized by using a Latin hypercube sampling (LHS) method to uniformly sample the hyperparameter values from the search space for the first population of architectures. These architectures are trained using the backpropagation algorithm and the objective functions from Equation (1) and (2) calculated. This initial set of solutions and objective function values is used to populate the set of non-dominated solutions (*NDS*) and the surrogate training population (*STP*) for the Random Forest fitting.

**Learning phase:** During the initial $LG$ generations, each subproblem $i \in \{1,..,N\}$ is solved once in a generation by randomly selecting two solutions from the neighborhood of $i$ and applying crossover and random mutation to generate a new solution. The architecture is trained and the objective functions calculated. This is used to update the population of neighboring solution (*PNS*) with the Penalty-based Boundary Intersection approach (PBI) [14], *NDS* and *STP*.

**Exploitation phase:** After the *LG* generations, the information produced in phases 1 and 2 are used to guide the search. In each generation, *N* subproblems are solved by selecting a subproblem according to the probability defined in [14]. For the selected subproblem, two solutions are randomly chosen from the neighborhood and a new solution generated by applying crossover and mutation operators. The hyperparameter value to mutate is chosen using the probability presented in Equation (5). The Random Forest surrogate is then trained with the STP population to predict the network´s ESE value and calculate the prediction´s dispersion. The new solution is selected to be trained if any of these 4 criteria is met: 1) the solution updates the *PNS* using the PBI approach, 2) the solution is predicted to be part of *NDS,* 3) the predicted ESE has the minimum value in the generation, or 4) the prediction´s dispersion has the highest value in the generation. If none of the criteria is satisfied, the solution is discarded and the algorithm identifies a new subproblem to solve. If the solution was selected to be trained, the true objective functions are calculated and used to update the *PNS, NSD* and *STP*. We only use solutions that have been trained to update the populations and selection probabilities to ensure the search is correctly guided and the Pareto Front truly approximated. After *G* generations, the *NDS* is returned.

In order to obtain a high quality solution, the search algorithm must find a balance between exploration and exploitation. During the initial LG generations (initialization and learning phase), the algorithm solves each subproblem once and sets an equal probability to all the hyperparameter values ($P_{h_{ij},G}$ has a uniform distribution). In this way, the algorithm starts by exploring all the hyperparameter search space and discovers how each subproblem contributes to the approximation of the Pareto Front. After the *LG* generations, an exploitation strategy is preferred. Therefore, the selection probabilities are computed and applied to guide the search towards the most promising hyperparameter values and subproblems. We have tested the proposed algorithm with *LG* = 8, 10, 12, 14 and 16 and *LG*=10 was selected because it improved the convergence speed without degrading the performance of the best solution found.

## 3. RESULTS

### 3.1 Implementation Results

EMONAS is evaluated on cardiac MR image segmentation from the MICCAI ACDC challenge [11]. The dataset is composed of 4D cine-MR images from 150 patients, from which 100 images with their ground truth segmentation are provided for training and 50 unlabeled images for testing. The task is the segmentation of the right ventricle cavity (RVC), left ventricle cavity (LVC), and left ventricle myocardium (LVM). The experiments are carried on one 8-GB GTX 1070Ti GPU.

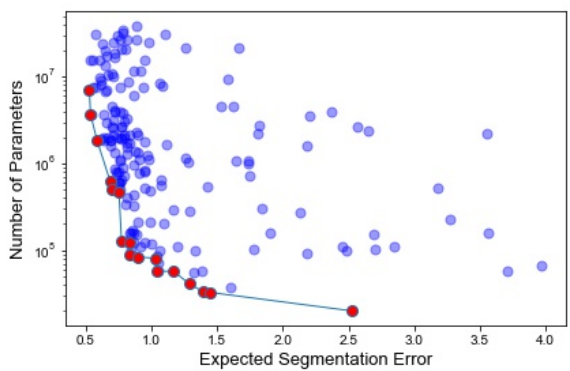

Figure 2. Pareto Front (red points) obtained with the SaMEA algorithm and trained candidate architectures (blue points) during evolution.

The training dataset is split into 80 training images and 20 validation images. The algorithm runs for a total of 40 generations, with a population size of 10, neighborhood size of 4 and $LG$=10. An $\alpha$ =0.25 and $\beta$=0.10 are applied for the ESE function. Additionally, the candidate architectures are partially trained for 60 epochs using patches of size 144×144×10 and the Adam optimizer. The number of training epochs has been set after testing it is able to satisfactory distinguish the quality of the constructed architectures. For the random forest, the number of regression trees is set to 100 and the minimum number of data points for split to 5 after using a random search method for the hyperparameter selection. The SaMEA algorithm runs for 11.43 days and obtains 17 non-dominated points. The Pareto Front obtained is shown in Figure 2.

Since the main objective of this problem is to produce an accurate segmentation, the solution with the smallest ESE function is selected as the best. Nevertheless, by incorporating the size of the model as a second objective function, models that have a small size are included in the population of neighboring solutions during evolution. Therefore, the candidate architectures that are constructed from the subproblems that prioritize the accuracy inherit the hyperparameters from the smaller architectures making them both efficient and accurate. It must be stated that providing the whole Pareto Front allows the researcher to select the most appropriate architecture for the accuracy and hardware constraints in their specific problem. The selected best architecture has $7.1\times10^6$ number of parameters. The best architecture is trained for 2000 epochs with patches of size 144×144×10, using the Adam optimizer and data augmentation. Furthermore, a largest connected component analysis is applied as post-processing operation. In Figure 3, the qualitative results of the EMONAS architecture on the validation set are presented. The segmentations of the three sub structures have an accurate shape and smooth contours.

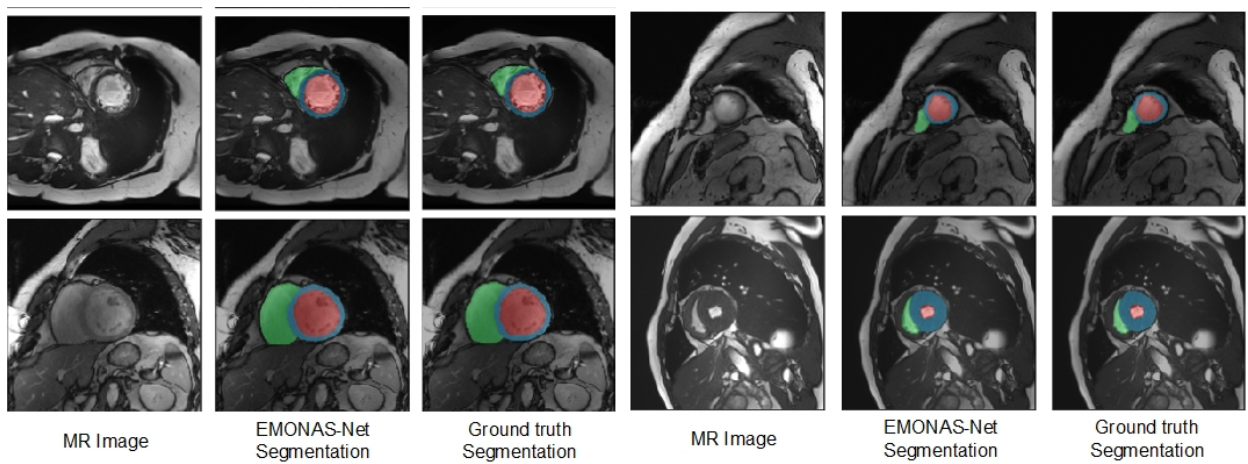

Figure 3. Example of the segmentation results of the EMONAS architecture. The red region denotes the LVC, the blue region the LVM and the green region the RVC.

### 3.2 Comparison with State-of-the-art Models

The best architecture is used for segmenting the 50 testing images and the evaluation was carried via an online submission to the MICCAI ACDC challenge. Geometrical and clinical performance measures are calculated to assess the segmentation. Due to space limitations, only the mean Dice similarity coefficient (DSC) and Hausdorff distance (HD) for the top performing groups in the challenge as of June 2020 are shown in Table 2. In the submissions, two groups applied an AutoML method to automatically design the architecture while the rest manually designed the networks for this specific task. This is denoted as design type in the tables. Considering that the competing AutoML methods use an ensemble of 10 networks for the final segmentation, we present the results using a single-model (EMONAS-1) and a 5-network ensemble (EMONAS-5) where the best architecture is trained using a 5-fold cross-validation setting.

Table 2. Comparison of EMONAS with top competing methods on the ACDC challenge test set.

| Group | Method | RVC DSC | RVC HD | LVC DSC | LVC HD | LVM DSC | LVM HD | Num. of Parameters |
|---|---|---|---|---|---|---|---|---|
| EMONAS-1 | AutoML | 91.24 | 11.34 | 92.59 | 7.81 | 87.59 | 8.72 | $7.1 \times 10^6$ |
| EMONAS-5 | AutoML | 91.20 | 11.18 | 93.10 | 7.35 | 88.00 | 8.92 | $35.5 \times 10^6$ |
| Isensee | AutoML | **92.75** | **9.93** | **94.75** | **6.20** | **91.35** | **7.17** | $235.5 \times 10^6$ |
| Zotti I | Manual | 91.15 | 12.19 | 93.80 | 7.28 | 89.40 | 9.44 | - |
| Zotti II | Manual | 90.95 | 11.85 | 93.10 | 7.67 | 89.00 | 8.99 | - |
| Painchaud | Manual | 90.85 | 13.52 | 93.60 | 7.22 | 88.90 | 9.12 | - |
| Baumgartner | Manual | 90.75 | 13.68 | 93.70 | 7.85 | 89.65 | 9.67 | - |
| Khened | Manual | 90.70 | 13.96 | 94.05 | 8.55 | 89.35 | 11.21 | - |
| Wolterink | Manual | 90.00 | 12.64 | 93.95 | 8.56 | 88.45 | 10.90 | - |
| Rohé | Manual | 88.05 | 14.99 | 92.85 | 9.12 | 86.80 | 12.29 | - |
| Jain | Manual | 86.50 | 16.12 | 92.00 | 9.57 | 88.95 | 10.51 | - |
| Grinias | Manual | 82.70 | 21.65 | 89.80 | 10.92 | 79.15 | 13.43 | - |
| Yang | Manual | 77.95 | 30.69 | 81.95 | 50.46 | - | - | - |

EMONAS has a competitive performance, being ranked within the top 10 submissions of the leaderboard in all evaluation metrics. In the segmentation of the RVC, EMONAS-5 is ranked second in the mean DSC and mean HD. EMONAS-5 is ranked third in terms of the mean HD for the LVM segmentation, and fourth in the mean HD for the LVC segmentation. Although EMONAS-1 has a slight statistically significant decrease in performance in terms of the LVC DSC, LVM DSC and LVC HD, it is still comparable to the other AutoML methods while being considerably smaller, and surpassing many manually designed architectures. As shown in Table 3, EMONAS-1 provides a reduction in the number of parameters of 33× against the model from Isensee et al. [25] and 3.4× against the model from Baldeon et al. [22].

### 3.3 Efficiency Evaluation

To evaluate the efficiency of EMONAS against other NAS methods, we compare it to a reinforcement learning (RL) based framework [4] tested on the ACDC dataset and the MEA algorithm [5] with the proposed search space as shown in Table 3. The MEA algorithm is implemented using a similar GPU as EMONAS, while [4] uses 15 Titan X GPUs. EMONAS leads the performance in most evaluation metrics in both segmentation accuracy and efficiency of the search. Against the RL framework, EMONAS finds an architecture with 4.2× fewer parameters. In relation to the MEA algorithm, the search time decreases by 52%.

Table 3. Comparison of EMONAS with competing NAS methods on the ACDC challenge test set.

| NAS Method | RVC DSC | RVC HD | LVC DSC | LVC HD | LVM DSC | LVM HD | Num. of Parameters | GPU days |
|---|---|---|---|---|---|---|---|---|
| EMONAS-1 | **91.2** | **11.3** | 92.6 | 7.8 | **87.6** | **8.7** | $7.1 \times 10^6$ | **11.4** |
| MEA [5] | 91.0 | 11.6 | 92.7 | **7.7** | 87.5 | 9.4 | $\mathbf{7.0 \times 10^6}$ | 23.8 |
| RL based NAS [4] | 86.8 | 14.3 | **92.8** | 8.9 | 84.9 | 10.7 | $30 \times 10^6$ | - |

## 4. CONCLUSIONS

In this work, we present EMONAS, an efficient multiobjective neural architecture search framework that optimizes the network's accuracy and size. EMONAS is composed of a novel micro and macro search space and a surrogate-assisted multiobjective evolutionary based algorithm (SaMEA). The proposed search space allows the joint optimization of the macro- and micro-structure of the architecture, which reduces the need for manual intervention. Meanwhile, the SaMEA algorithm implements selection probabilities and a Random Forest surrogate to efficiently explore the search space and decrease the search time. EMONAS was evaluated on the task of 3D cardiac segmentation. The experiments demonstrate EMONAS can automatically find highly competitive and efficient architectures while reducing the search time by more than 50%.